\documentclass[letterpaper, 10 pt, conference]{ieeeconf}  

\IEEEoverridecommandlockouts                              

\usepackage{graphics} 
\usepackage{epsfig} 
\usepackage{times} 
\usepackage{amsmath,amssymb,amsfonts}
\usepackage{newtxtext, newtxmath}
\usepackage{autobreak}
\usepackage{booktabs}
\usepackage{caption}
\usepackage{cite}
\usepackage{xcolor}
\usepackage{marvosym}

\title{\LARGE \bf
CRAFT: Adapting VLA Models to Contact-rich Manipulation via Force-aware Curriculum Fine-tuning
}

\author{Yike Zhang$^{1,2}$, Yaonan Wang$^{1}$, Xinxin Sun$^{1,2}$, Kaizhen Huang$^{1,2}$, Zhiyuan Xu$^{2}$, \\
Junjie Ji$^{2,\dag}$, Zhengping Che$^{2,\dag}$, Jian Tang$^{2}$, Jingtao Sun$^{1,3,}$\textsuperscript{\Letter}
\thanks{
  This work was in part funded and supported by Beijing Innovation Center of Humanoid Robotics.
  $^{\dag}$Project Leads: Junjie Ji and Zhengping Che.
  \textsuperscript{\Letter}Corresponding Author: Jingtao Sun.
}
\thanks{$^{1}$National Engineering Research Center of Robot Visual Perception and Control Technology, Hunan University, Changsha, 410082, China.
        {Yike Zhang, Yaonan Wang, Kaizhen Huang, and Xinxin Sun}: {\tt\small \{b240900525, yaonan, huangkaizhen, s2001\}@hnu.edu.cn}}
\thanks{$^{2}$Beijing Innovation Center of Humanoid Robotics, Beijing, 102600, China.
        {Zhiyuan Xu, Junjie Ji, Zhengping Che, and Jian Tang}: {\tt\small \{eric.xu, jacob.ji, z.che, jian.tang\}@x-humanoid.com}}
\thanks{$^{3}$Department of Electrical and Computer Engineering, National University of Singapore, 119077, Singapore.
        Jingtao Sun: {\tt\small jingtao\_sun@nus.edu.sg}}%
}

\begin{document}

\maketitle
\thispagestyle{empty}
\pagestyle{empty}

\begin{abstract}
Vision–Language–Action (VLA) models have shown a strong capability in enabling robots to execute general instructions, yet they struggle with contact-rich manipulation tasks, where success requires precise alignment, stable contact maintenance, and effective handling of deformable objects. A fundamental challenge arises from the imbalance between high-entropy vision and language inputs and low-entropy but critical force signals, which often leads to over-reliance on perception and unstable control. To address this, we introduce \textbf{CRAFT}, a force-aware curriculum fine-tuning framework that integrates a variational information bottleneck module to regulate vision and language embeddings during early training. This curriculum strategy encourages the model to prioritize force signals initially, before progressively restoring access to the full multimodal information. To enable force-aware learning, we further design a homologous leader–follower teleoperation system that collects synchronized vision, language, and force data across diverse contact-rich tasks. Real-world experiments demonstrate that CRAFT consistently improves task success, generalizes to unseen objects and novel task variations, and adapts effectively across diverse VLA architectures, enabling robust and generalizable contact-rich manipulation.
\end{abstract}

\section{Introduction}

The emergence of large-scale foundation models, including Vision-Language Models (VLMs) and Large Language Models (LLMs), has significantly advanced artificial intelligence and robotics. These models exhibit strong capabilities in perception, reasoning, and grounding, enabling robots to follow natural language instructions and adapt across diverse environments. By leveraging massive pretraining and multimodal alignment, foundation models have become a promising paradigm for scalable robot intelligence.

With the advent of powerful foundation models, the Vision–Language–Action (VLA) paradigm has emerged as a promising framework that integrates pretrained visual and language representations with robot learning systems, translating high-level semantic goals into low-level motor actions. Recent works such as RT-1/RT-2~\cite{brohan2023rt,zitkovich2023rt}, SayCan~\cite{ahn2022can}, and OpenVLA~\cite{kim2025openvla} demonstrate strong zero-shot generalization to novel objects, instructions, and environments, marking an important milestone for general-purpose robotic agents. More recent approaches, including RDT~\cite{liurdt} and $\pi_0$~\cite{black2024pi_0}, further underscore the potential of unifying large-scale multimodal pretraining with robot control, highlighting scalability across diverse tasks and robot architectures.

However, despite recent advances, contact-rich manipulation remains a fundamental challenge for VLA models. Tasks such as precise insertions, deformable object rolling, or continuous surface wiping require precise alignment and dynamic regulation of contact forces, yet existing models often fail to meet these demands. This limitation arises from two key issues. First, relying predominantly on high-entropy visual and language inputs makes it difficult to achieve the fine alignment required in contact-rich settings, as perceptual inputs alone cannot capture subtle physical interactions at the point of contact. Second, the absence of explicit force or tactile feedback deprives models of low-entropy but task-critical signals that reveal resistance or slip, leading to unstable, imprecise, or even unsafe behaviors. 

\begin{figure}[t]
    \centering
    \includegraphics[width=0.95\linewidth]{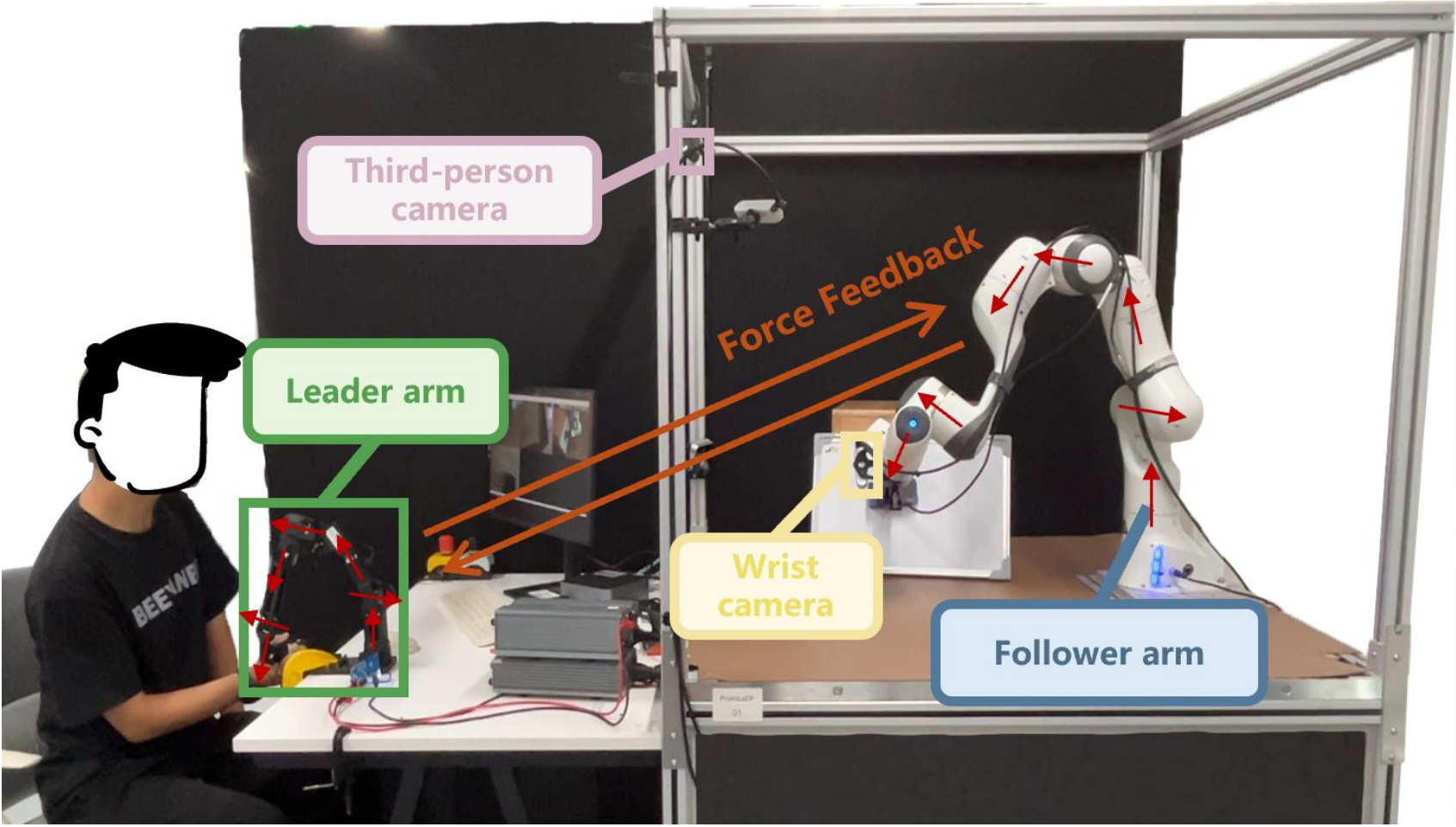}
    \caption{\textbf{Homologous leader–follower teleoperation system.} The operator controls the leader arm while the follower arm mirrors the motion. Real-time force feedback enables natural interaction sensing, and two synchronized cameras capture first-person and third-person visual observations.}
    \label{fig:teleoperation_system}
    \vspace{-0.5cm}
\end{figure}

Several prior works have sought to improve contact-rich manipulation. Reactive Diffusion Policy~(RDP)~\cite{xue2025reactive} introduces a slow–fast visual–tactile policy learning scheme to improve stability and responsiveness in contact-rich manipulation. VLA-Touch~\cite{bi2025vla} enhances VLA models with dual-level tactile feedback, demonstrating the importance of incorporating fine-grained touch signals alongside visual and language inputs. ForceVLA~\cite{yu2025forcevla} further extends this line of work by introducing a force-aware mixture-of-experts architecture, enabling VLA models to better exploit force information in physically demanding tasks. 
While these methods represent promising steps toward force-aware VLA models, they remain limited by modality entropy imbalance and the lack of a unified mechanism to regulate multimodal contributions. As a result, current approaches still struggle to prioritize low-entropy yet critical force signals, leaving robust and generalizable contact-rich manipulation as an open challenge.



To address these challenges, we present \textbf{CRAFT}, a lightweight and model-agnostic framework for adapting pretrained VLA models to contact-rich manipulation. CRAFT introduces a variational information bottleneck (VIB)~\cite{alemi2017deep} after the vision–language encoder to temporarily restrict the information flow from visual and language embeddings. Within a curriculum fine-tuning scheme, the VIB initially imposes a strong  Kullback–Leibler~(KL) penalty that suppresses high-entropy visual and language signals, compelling the policy to focus on low-entropy force inputs and thereby acquire stable, contact-aware behaviors. As training progresses, the weight of the VIB regularization is gradually annealed, releasing the bottleneck and allowing the model to reincorporate rich perceptual information while retaining the previously learned force-sensitive representations. This staged force-first-then-multimodal strategy enables CRAFT to efficiently and seamlessly adapt VLA models to contact-rich manipulation tasks, while preserving their multimodal perception capabilities.

We empirically validate CRAFT on five challenging real-world tasks, including fine-alignment insertions, continuous contact wiping, and manipulation of deformable objects. Experimental results demonstrate that CRAFT improves task success rates, enhances generalization to unseen objects and novel tasks, and adapts to multiple popular VLA architectures such as RDT and $\pi_0$. These findings highlight the effectiveness of combining information-theoretic modality control with force-aware curriculum fine-tuning to enable robust, generalizable contact-rich manipulation in VLA models. Overall, our main contributions are summarized as follows:
\begin{itemize}
    \item We introduce \textbf{CRAFT}, a general framework that adapts VLA models to \textbf{C}ontact-\textbf{R}ich manipulation with a Force-\textbf{A}ware curriculum \textbf{F}ine-\textbf{T}uning strategy, leveraging a variational information bottleneck to balance the influence of vision, language, and force modalities.
    \item We develop a homologous leader-follower teleoperation system for data collection based on an off-the-shelf robot arm without additional sensor installation, which enables synchronized vision, language, and force recording across diverse contact-rich tasks, which facilitates robust policy learning and evaluation of force-aware VLA models.
    \item Extensive real-world experiments demonstrate that our proposed CRAFT enhances task success, improves generalization to unseen objects and novel tasks, and is broadly applicable across different VLA-based approaches.
\end{itemize} 
\section{RELATED WORK}

\subsection{Vision-Language-Action Models}
VLA models aim to map visual observations and high-level language commands into robot actions. Early works primarily focused on learning vision-to-action mappings from large-scale real-world datasets~\cite{jiang2022vima,zhao2023learning,huang2023voxposer,chi2023diffusion,team2024octo,11127918}, often relying on extensive demonstrations and task-specific engineering. While effective for individual tasks, these approaches typically struggle to generalize to novel objects or instructions. More recent methods incorporate semantic priors, generative modeling, or diffusion-based action prediction to enhance zero-shot generalization~\cite{yue2024deer,dey2024revla,wei2024occllama,li2024cogact,sun2024l4d,wen2025dexvla,budzianowski2025edgevla,li2025switchvla,sun2025towards,su2025freqpolicy}. Recent representative approaches~\cite{liurdt,black2024pi_0} leverage diffusion or flow-based frameworks to predict complex action sequences, enabling more flexible and scalable policy representations. These models demonstrate the potential of unifying pretrained visual and language representations with robot control, allowing for the execution of diverse manipulation tasks from high-level instructions without task-specific retraining.

Despite these advances, most VLA models largely ignore force or tactile feedback~\cite{thomaz2008teachable,weinberg2024survey,zhao2025polytouch}, which are critical for contact-rich manipulation tasks such as precise insertions, rolling deformable objects, or continuous surface wiping. Several recent works have attempted to incorporate force or tactile signals as auxiliary inputs. For example, RDP introduces a slow-fast visual-tactile policy learning scheme to improve responsiveness and stability~\cite{xue2025reactive}, VLA-Touch incorporates dual-level tactile feedback to complement visual and language modalities~\cite{bi2025vla}, and ForceVLA employs a force-aware mixture-of-experts architecture to leverage force signals in predicting action sequences~\cite{yu2025forcevla}. Closely related to our work, FACTR~\cite{liu2025factr} proposes a curriculum learning strategy that corrupts visual inputs with decreasing intensity via Gaussian blur to prioritize force signals early in training. However, while FACTR relies on heuristic schedules of input degradation to shift modality focus, CRAFT leverages the Information Bottleneck principle~\cite{tishby2000information} to explicitly compress high-entropy visual and language representations. This provides a theoretically grounded mechanism to regulate multimodal information flow, ensuring that force features are robustly established in the latent space before fully reintegrating dense perceptual data. As a result, scalable and generalizable force-aware VLA learning remains an open challenge, motivating the need for a framework that systematically integrates and balances multimodal signals for robust contact-rich manipulation.

\subsection{Information Bottleneck in Multimodal Learning}
The Information Bottleneck~(IB)~\cite{tishby2000information} provides a principled framework to compress input redundancy while preserving task-relevant information, balancing representation compactness with predictive performance. Its variational extension, the deep variational information bottleneck~(VIB)~\cite{alemi2017deep}, enables efficient implementation in deep neural networks by approximating the IB objective with tractable variational distributions, and has been widely adopted in multimodal and high-dimensional representation learning for its ability to reduce noise, prevent overfitting, and improve interpretability.

In multimodal settings, IB-based approaches have been applied to align heterogeneous modalities such as image and text. Graph IB~\cite{wu2020graph} and Learning with Decodable IB~\cite{dubois2020learning} focus on learning compact, task-relevant embeddings while preserving decodability, and Restricting the Flow~\cite{schulz2020restricting} demonstrates IB's utility for controlling information flow in attribution tasks. Similarly, the Deep Copula IB~\cite{wieczorek2018learning} learns sparse latent representations that capture essential cross-modal dependencies. More recent work, such as the Dynamic Multimodal Information Bottleneck~\cite{fang2024dynamic}, introduces adaptive compression–fidelity constraints that allow dynamic control over each modality’s contribution in fused representations. These techniques have been used to improve robustness, interpretability, and generalization in models such as CLIP~\cite{radford2021learning}.

\begin{figure*}[t]
    \centering
    \includegraphics[width=0.95\linewidth]{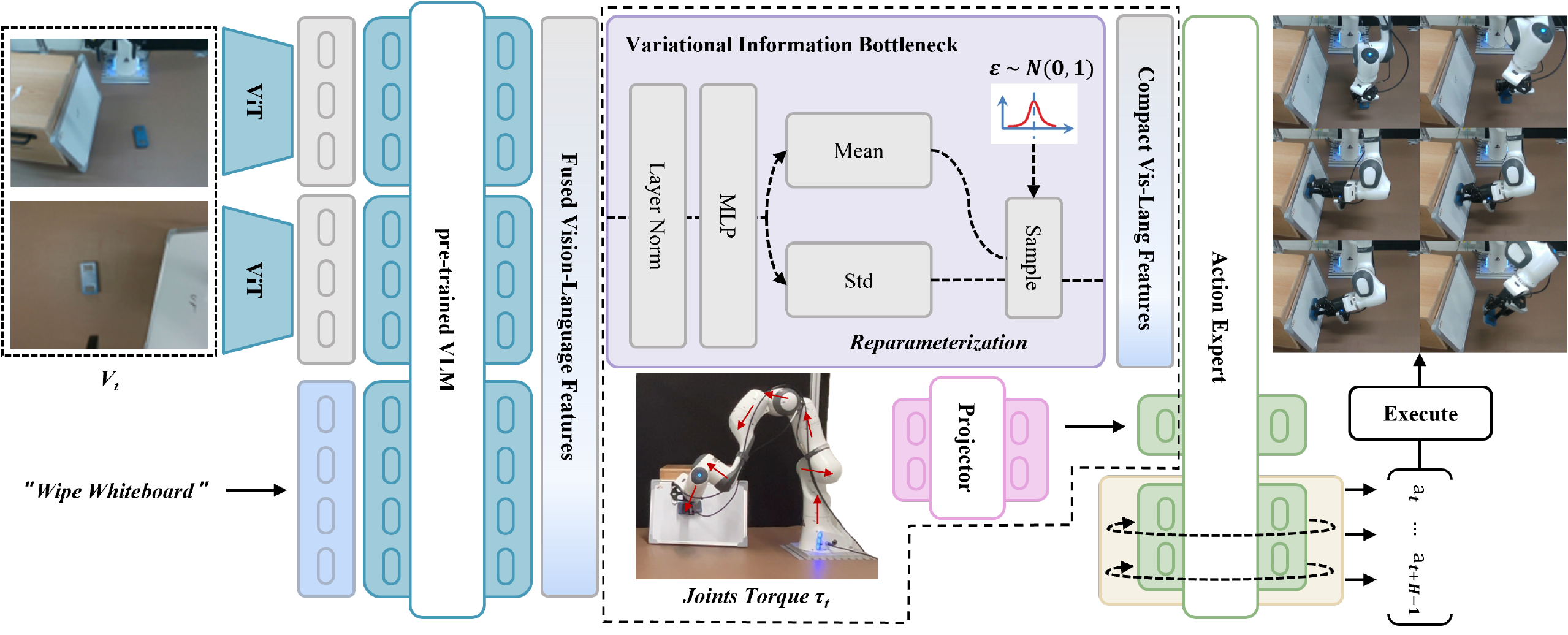}
    \caption{\textbf{Overview of CRAFT.} CRAFT mitigates the imbalance between high-entropy perceptual signals (vision and language) and low-entropy force signals in contact-rich manipulation. The VIB module initially compresses visual and language features to emphasize force information, and a curriculum weight schedule gradually relaxes this constraint, allowing the policy to integrate rich multimodal signals while retaining force-aware representations. CRAFT can be applied to diverse VLA models.}
    \label{fig:pipeline}
\end{figure*}

Despite these advances, most IB-based methods concentrate on perception-level representations, with limited efforts directed toward regulating the relative influence of vision, language, and force modalities in VLA frameworks for contact-rich manipulation. Existing attempts to integrate tactile or force signals into policy learning often remain at the level of auxiliary features or low-level controllers~\cite{wang2023visual,xie2025towards}, lacking mechanisms to adaptively balance multimodal contributions during action prediction. Consequently, there remains a significant gap in developing scalable and generalizable force-aware VLA models that can systematically balance multimodal information throughout both learning and execution, motivating our proposed CRAFT framework that leverages VIB to regulate vision, language, and force contributions for robust contact-rich manipulation. 
\section{Method}

While visual and language signals provide high-entropy information about the scene, force signals are inherently lower in entropy but uniquely capture direct physical interactions, making them indispensable for stable and precise contact-rich manipulation.
In this section, we introduce CRAFT, a general framework that equips existing VLA models with the ability to perform contact-rich manipulation guided by force information, as shown in Fig.~\ref{fig:pipeline}.
CRAFT inserts a variational information bottleneck after vision and language embeddings to regulate their information entropy, enforcing a curriculum that encourages the model to prioritize the force modality during the early stages of training, and gradually restores the contribution of visual and language signals thereafter.
\subsection{Homologous Leader-Follower Teleoperation System}
To enable force-aware curriculum fine-tuning, we design a homologous leader–follower teleoperation system inspired by HACTS~\cite{xu2025hacts} and FACTR~\cite{liu2025factr}. Our setup employs an off-the-shelf robot arm, i.e., Franka Emika Panda, without the need for additional tactile or force sensors, alongside a leader arm constructed from low-cost servo motors with built-in force control, eliminating the reliance on high-precision and expensive haptic equipment such as HaptX Gloves or specialized haptic devices. As illustrated in Fig.~\ref{fig:teleoperation_system}, the operator controls the leader arm while the follower arm precisely mirrors the commanded motions. Contact forces measured at the follower are fed back to the leader in real time, allowing the operator to naturally perceive interaction forces and provide high-quality demonstrations. The bidirectional synchronization in our system enables fine-grained force control during teleoperation, ensuring that the collected force data is both accurate and reliable. 

The robot operates under impedance control, which modulates compliance during physical interaction to prevent mechanical locking and ensure safe, stable contact. Joint torque is used as the force modality, providing richer and more informative interaction data than joint positions alone, while remaining broadly accessible across most robotic platforms.

Visual observations are captured from two synchronized cameras: a wrist-mounted camera for first-person perspectives and a static third-person camera for environmental context. 
All sensory modalities—including multi-view images, proprioceptive states, joint torques, and robot actions—are temporally synchronized and processed for seamless integration into the CRAFT framework, providing a comprehensive multimodal foundation for force-aware policy learning.

\subsection{Force-aware Curriculum Fine-tuning}
Directly incorporating force modality alongside visual and language inputs often results in the force modality being underutilized, especially given the sparsity and lower information content of the force signal compared to rich visual and language signals. Consequently, the model tends to overly rely on visual and language signals, limiting its ability to leverage critical contact information encoded in the force modality.

To mitigate this imbalance, we propose Force-aware Curriculum Fine-tuning, a strategy that initially emphasizes the force modality by constraining the information from visual and language signals through a variational information bottleneck module with KL divergence regularization. This constraint is gradually relaxed via a weight schedule, enabling the model to progressively reintegrate rich multimodal information while retaining the force-aware representations learned during early training. Notably, this strategy can be applied as a plug-in to various pretrained VLA models, providing a unified framework that equips them with force-aware capabilities.

\noindent\textbf{Variational Information Bottleneck (VIB).}
In contact-rich manipulation, the force signal plays a crucial role in stabilizing the interaction and detecting external constraints. However, considering the multimodal policies that jointly process vision, language, and proprioception inputs, the high entropy of visual and language features often dominates the representation space, making it difficult for the policy to exploit force-related signals effectively. To address this, we applied the VIB~\cite{alemi2017deep} module to the visual and language embeddings $F_{V}$ and $F_{L}$, encouraging the model to prioritize low-entropy, task-relevant proprioception.


As illustrated in Fig.~\ref{fig:pipeline}, at each timestep $t$, the policy observes $O_t = \{V_t, \tau_t\}$ and receives a language instruction $L$, producing a sequence of actions $A_t = \{a_t, \dots, a_{t+H-1}\}$ over a horizon $H$. Since visual and language signals typically exhibit significantly higher entropy $\mathcal{H}(V_t), \mathcal{H}(L) \gg \mathcal{H}(\tau_t)$, where $\mathcal{H}(\cdot)$ denotes entropy, a naive policy tends to overfit to perceptual modalities while ignoring the force modality $\tau_t$.

To mitigate this imbalance, we compress the fused visual and language embeddings $F_{V_t}$ and $F_L$ into compact latent features $F^c_{V_t}$ and $F^c_L$ via the VIB module, and minimize the mutual information between the original and compact features:
\begin{equation}
    I(F, F^c) = \mathbb{E}_{p(f,f^c)} \left[ \log \frac{p(f,f^c)}{p(f)p(f^c)} \right].
\end{equation}

Since the true marginal $p(F^c)$ is intractable, we introduce a variational approximation $q(F^c)$, which yields the following upper bound:
\begin{equation}
    \mathcal{L}_{\text{VIB}} \leq \mathbb{E}_{p(f)} \big[ KL( p(F^c|F) \,\|\, q(F^c) ) \big],
\end{equation}
where $KL(\cdot)$ represents the Kullback-Leibler divergence loss~\cite{kullback1951information}. We set the prior $q(F^c)$ to a standard normal $\mathcal{N}(0, 1)$ and model the posterior as $p(F^c | F) = \mathcal{N}(\mu, \sigma^2)$, where $\mu$ and $\sigma$ are predicted by separate linear layers. This leads to the VIB loss:
\begin{equation}
    \mathcal{L}_{\text{VIB}} = \frac{1}{2} \sum_i \left( -\log \sigma_i^2 + \mu_i^2 + \sigma_i^2 - 1 \right).
\end{equation}

At the stage of fine-tuning, the compact features are sampled via the reparameterization trick:
\begin{equation}
    F^c = \mu + \sigma \cdot \epsilon, \quad \epsilon \sim \mathcal{N}(0, I).
\end{equation}

This construction dynamically regulates the information from high-entropy perceptual modalities. During the early stages of finetuning, KL regularization encourages the policy to rely more heavily on low-entropy proprioception features. As training advances, the VIB gradually relaxes, allowing vision and language to contribute complementary information once the force-aware policy is stabilized.

\noindent\textbf{Force as Proprioception.}
In robotic manipulation, particularly for contact-rich tasks, it is crucial for the policy to accurately perceive the robot-environment interaction forces. Rather than relying on joint positions $q$ as the proprioception input, we leverage joint torques $\tau$, which inherently capture richer force-aware information and serve as a task-relevant representation. Under standard joint-level impedance control, the commanded torque can be expressed as:
\begin{equation}
    \tau = K(q_d - q) + D(\dot q_d - \dot q) + M(\ddot q_d - \ddot q) + \tau_{\text{ext}},
\end{equation}
where $q, \dot q, \ddot q$ denote the current joint positions, velocities, and accelerations, respectively,
$q_d, \dot q_d, \ddot q_d$ represent the desired trajectories, 
$K, D$ are the impedance gains, 
$M$ is the joint-space inertia matrix, 
and $\tau_{\text{ext}}$ captures the external torques arising from environment contacts.

This formulation reveals that torque $\tau$ encodes not only the current kinematic state $(q,\dot q,\ddot q)$, but also:
\begin{itemize}
    \item Task-level deviations $(q_d - q)$ and $(\dot q_d - \dot q)$, representing how far the system is from the desired motion.
    \item Dynamic discrepancies $(\ddot q_d - \ddot q)$, capturing inertial effects that are absent when using $q$ alone.
    \item External interaction forces $\tau_{\text{ext}}$, which are essential for understanding contacts and constraints.
\end{itemize}

From an information-theoretic perspective, torque signals generally carry higher information content than positions alone. Since $\tau$ is a deterministic function of multiple system variables, including desired and measured accelerations, its entropy satisfies:
\begin{equation}
    \mathcal{H}(\tau) = \mathcal{H}\big(f(q, \dot q, \ddot q, q_d, \dot q_d, \ddot q_d, \tau_{\text{ext}})\big) 
    \ge \mathcal{H}(q),
\end{equation}
where $f(\cdot)$ denotes the impedance control mapping. Consequently, the mutual information between $\tau$ and the underlying system dynamics satisfies:
\begin{equation}
    \begin{split}
        I(\tau;\,[q, \dot q, \ddot q, q_d, \dot q_d, \ddot q_d, \tau_{\text{ext}}])\\
        \ge I(q;\,[q, \dot q, \ddot q, q_d, \dot q_d, \ddot q_d, \tau_{\text{ext}}]) .      
    \end{split}
\end{equation}

By feeding joint torque into the policy alongside visual and language features, the model achieves improved observability over both internal system dynamics and external contacts, thereby facilitating the learning of robust, contact-aware manipulation strategies.

\noindent\textbf{VIB Weight Schedule.} 
To effectively integrate multimodal information while encouraging force-aware learning, we introduce a dynamic scheduling strategy for the VIB regularization term in the total loss.
At the beginning of training, we assign a relatively high weight $\lambda_{\text{VIB}}$ to the VIB regularization, which strongly compresses the high-entropy visual and language features.
This forces the policy to prioritize the low-entropy torque signals, enabling the model to acquire robust contact-aware representations in the early stages.
As training advances, $\lambda_{\text{VIB}}$ is gradually decreased according to an exponential decay schedule, allowing the model to progressively reintegrate the visual and linguistic information without losing the force-dominated representations learned previously.

Formally, the overall loss function is:
\begin{equation}
    \mathcal{L}_{\text{total}} = \mathcal{L}_{\text{task}} + \lambda_{\text{VIB}}(t) \cdot \mathcal{L}_{\text{VIB}},
\end{equation}
where $\mathcal{L}_{\text{task}}$ denotes the original imitation learning loss, and $\lambda_{\text{VIB}}(t)$ is the VIB weight at training step $t$. Furthermore, we define the VIB weight schedule as follows:
\begin{equation}
    \lambda_{\text{VIB}}(t) = \lambda_{\text{init}} \cdot \exp\left(-\frac{t}{T_{\text{decay}}}\right),
\end{equation}
where $\lambda_{\text{init}}$ is the initial VIB weight and $T_{\text{decay}}$ is a decay constant controlling the speed of information reintegration.
This exponential decay enforces strong information compression during the early training phase, ensuring the policy focuses on torque-based force signals, while progressively reducing the regularization to enable richer integration of visual and language modalities in later stages.

\subsection{Implementation}
To demonstrate the generalizability of CRAFT, we implement it on two representative VLA models, the flow-based $\pi_0$ and the diffusion-based RDT. 
For fine-tuning, we first collect force-aware demonstrations using our homologous leader-follower teleoperation system, providing high-quality and diverse data on various contact-rich tasks. The given VLA models (pretrained only with vision and language inputs) are then adapted to these tasks using their corresponding force-aware demonstrations. 
During fine-tuning, joint torques are introduced as proprioceptive inputs, and a lightweight VIB module is added after the pretrained vision–language encoder to regulate the information content of high-entropy visual and language modalities. This design enables CRAFT to effectively balance multimodal contributions during manipulation and can be seamlessly integrated into other VLA models without architectural modifications. Moreover, the VIB module is broadly applicable to low-entropy, task-critical modalities: while we focus on joint torques here, the same mechanism can be naturally extended to tactile feedback and similar proprioceptive inputs.

\section{Experiments}

We conduct real-world contact-rich manipulation experiments to evaluate the effectiveness of CRAFT in augmenting VLA models, guided by three key research questions:  

\begin{enumerate}
    \item \textbf{Main Results:} How does CRAFT enable VLA models to achieve superior performance in contact-rich manipulation tasks compared to their original counterparts?
    
    \item \textbf{Generalization Studies:} How do CRAFT-enhanced VLA models demonstrate stronger generalization to unseen objects and novel task variations?
    
    \item \textbf{Ablation Studies:} Why is the proposed Force-Aware Curriculum Finetuning design critical, and what is the contribution of each component to the overall effectiveness of CRAFT?  
\end{enumerate}

\begin{figure}[t]
    \centering
    \includegraphics[width=0.95\linewidth]{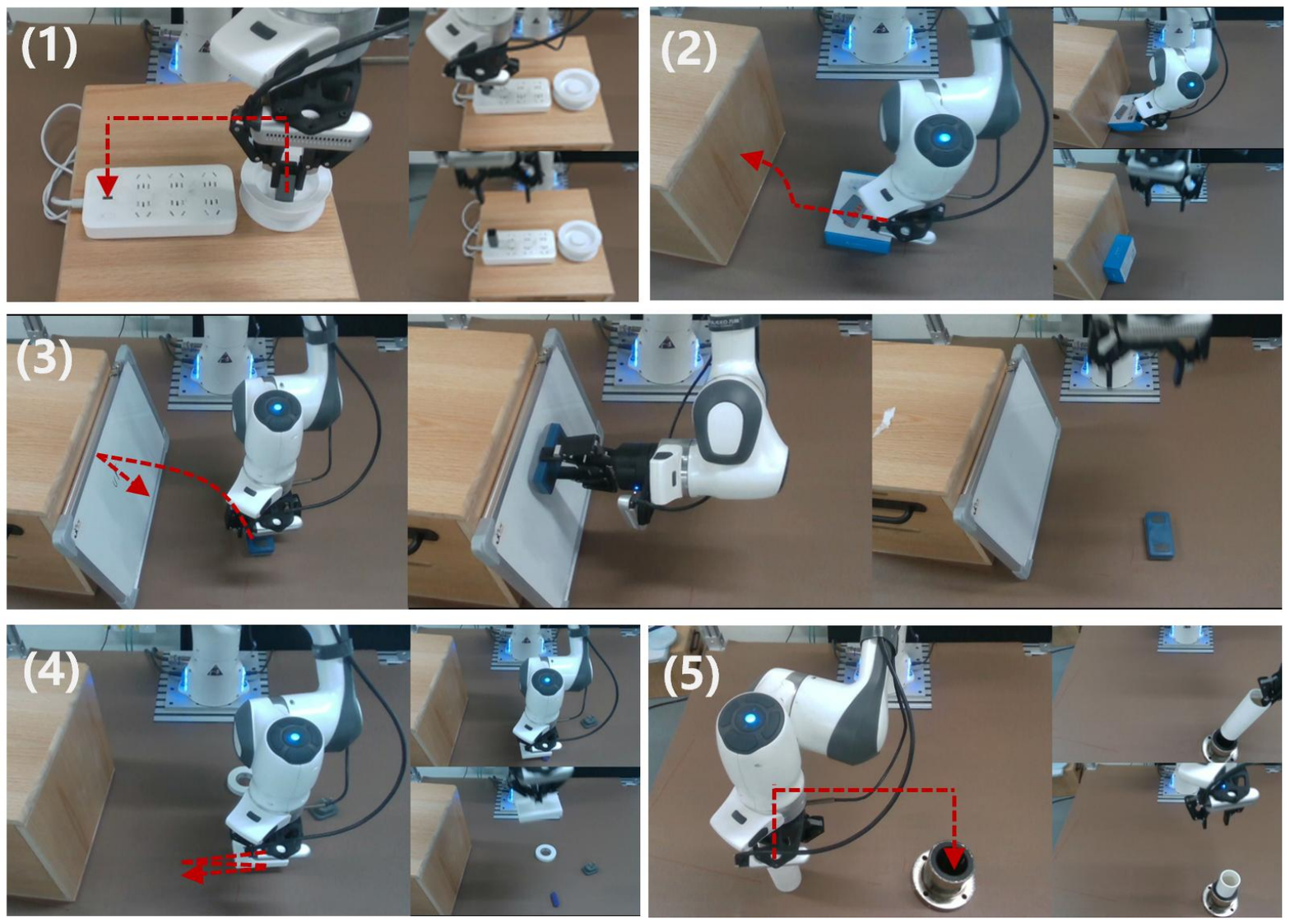}
    \caption{\textbf{Representative contact-rich tasks.} (1) USB Insertion, (2) Flip Carton, (3) Wipe Whiteboard, (4) Rolling Plasticine, and (5) Shaft-to-Hole Insertion.}
    \label{fig:task}
\end{figure}

\subsection{Experimental Setup}
\textbf{Tasks.} We consider five representative contact-rich tasks: USB Insertion, Flip Carton, Wipe Whiteboard, Rolling Plasticine, and Shaft-to-Hole Insertion, as shown in Fig.~\ref{fig:task}. These tasks involve precise alignment, continuous surface interaction, and manipulation of deformable objects. Task-specific success criteria are defined as follows:
\begin{itemize}
    \item USB/Shaft-to-Hole Insertion: The robot must insert a plug or shaft into a corresponding socket or hole, requiring precise alignment and controlled force application. A trial is considered successful if the plug or shaft is fully inserted into the corresponding socket or hole.
    \item Flip Carton: The robot must flip a carton from horizontal orientation to vertical orientation while maintaining stability. A trial is successful if the target carton is flipped over completely without dropping or damaging it.
    \item Wipe Whiteboard: The robot must wipe a designated area on a whiteboard with consistent contact. Success is determined by achieving at least 90\% uniform wiping coverage of the marked area.
    \item Rolling Plasticine: The robot must roll a block of plasticine into a cylindrical shape, requiring careful force control to avoid tearing. A trial is successful if the plasticine is rolled into a cylindrical shape without breaking or excessive deformation.
\end{itemize}

\begin{figure*}[t]
    \centering
    \includegraphics[width=0.95\linewidth]{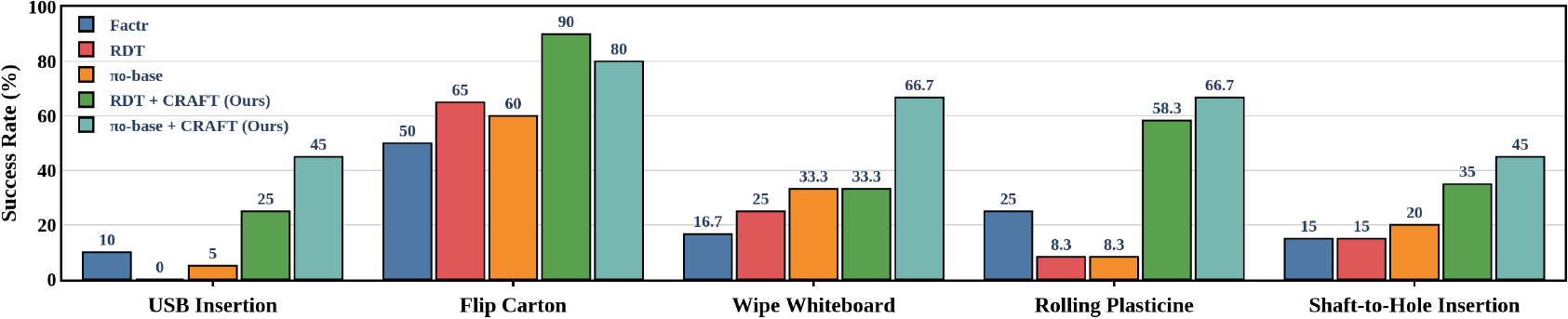}
    \caption{\textbf{Task success rates of baselines and CRAFT-enhanced VLA models.} Across five representative manipulation tasks, CRAFT consistently improves success rates for both $\pi_0$-base and RDT. The gains are particularly significant for contact-rich tasks, demonstrating the effectiveness of our force-aware curriculum fine-tuning strategy.}
    \label{fig:main_results}
\end{figure*}

\textbf{Training and Evaluation Protocol.} Each policy is trained with 50 teleoperated demonstrations per task, with randomized object poses to improve robustness. For evaluation, we perform a fixed number of trials for each task: 20 trials for USB Insertion, Shaft-to-Hole Insertion, and Flip Carton, and 12 trials for Wipe Whiteboard and Rolling Plasticine. 

\textbf{Baselines and Comparisons.} We compare CRAFT-enhanced models against representative VLAs and a force-aware baseline:

\begin{itemize}
    \item FACTR~\cite{liu2025factr}: A force-aware ACT~\cite{zhao2023learning} model, serving as a baseline for contact-rich manipulation tasks.  
    \item $\pi_0$-base~\cite{black2024pi_0}: A representative VLA model using flow matching but without force control.  
    \item RDT~\cite{liurdt}: Another representative VLA model based on DiT~\cite{peebles2023scalable} without force control.  
    \item $\pi_0$-base/RDT + CRAFT (Ours): $\pi_0$/RDT with force-aware curriculum fine-tuning.  
\end{itemize}

\subsection{Main Results}

\textbf{CRAFT leads to better performances.}
As shown in Fig.~\ref{fig:main_results}, our results demonstrate that the proposed CRAFT module significantly enhances the performance of VLA models in contact-rich manipulation tasks. Specifically, integrating CRAFT with the $\pi_0$-base model improves performance across five tasks, yielding an average increase of 35.36\%. Notably, for the Wipe Whiteboard task, CRAFT enables a remarkable improvement of 33.4\%(from 33.3\% to 66.7\%), underscoring its ability to enhance precision in tasks requiring sustained contact and force feedback. Similarly, RDT + CRAFT outperforms the baseline RDT model, advancing from 22.66\% to 48.32\%, with an average improvement of 25.66\% points and a particularly strong gain of 50\% in Rolling Plasticine (from 8.3\% to 58.3\%), highlighting CRAFT's effectiveness in tasks demanding dynamic force control. Compared to the available baselines, FACTR and both CRAFT-enhanced models achieve superior results, with $\pi_0$-base + CRAFT consistently attaining the highest success rate across most tasks. These findings validate that CRAFT leads to better performance by enabling effective force-aware control in VLA models.

\begin{figure}[t]
    \centering
    \includegraphics[width=0.95\linewidth]{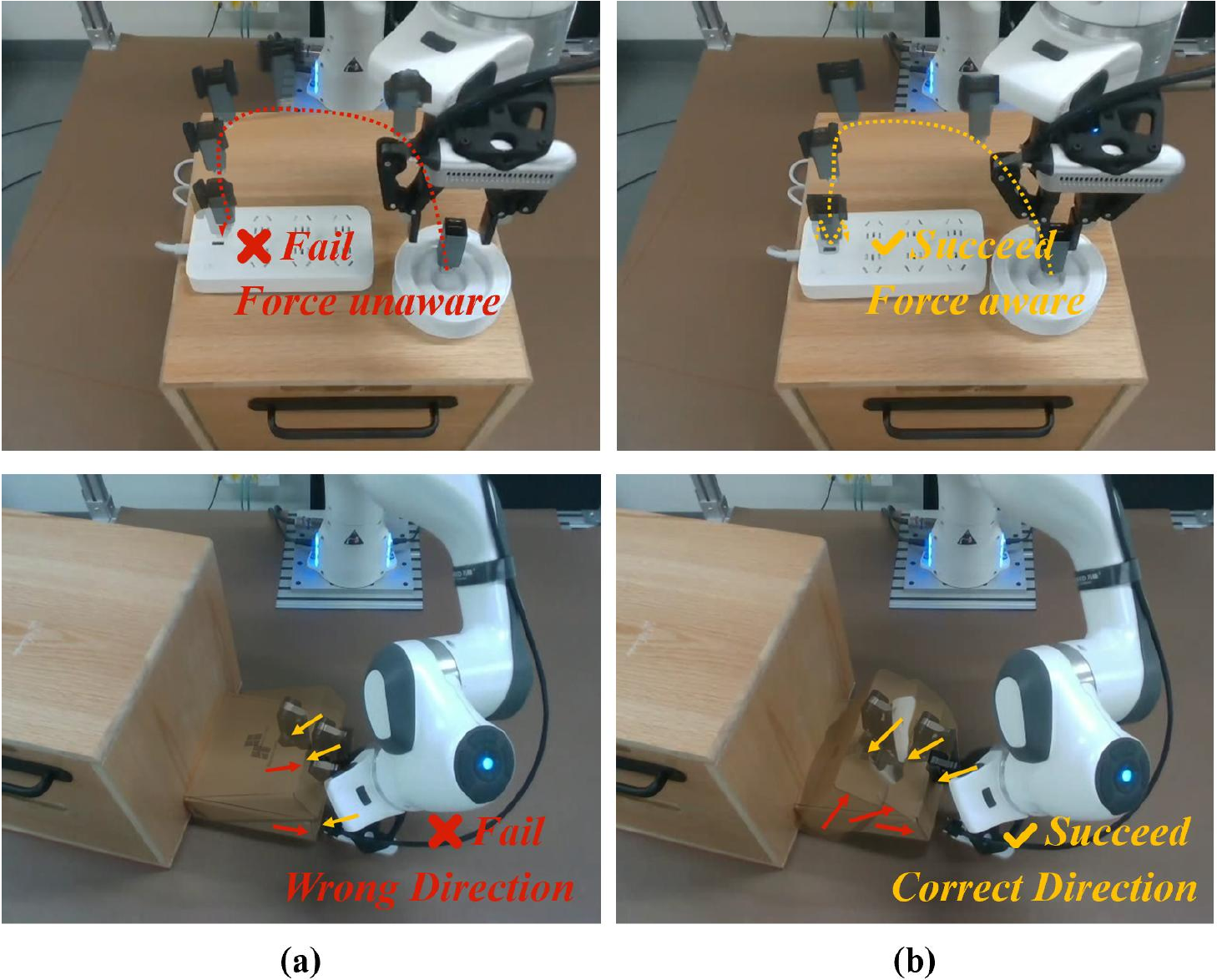}
    \caption{\textbf{CRAFT enables force-aware manipulation.} Comparison of $\pi_0$-base (a) and CRAFT-enhanced (b) on USB insertion and flip carton. CRAFT uses force signals to adjust motions, enabling successful execution.}
    \label{fig:force_aware}
\end{figure}

\textbf{CRAFT enables force-aware control.} 
By leveraging joint torque feedback, CRAFT-enhanced models adapt their manipulation strategies in real time, enabling precise alignment, stable contact maintenance, and safer physical interactions in contact-rich environments.
As shown in Fig.~\ref{fig:force_aware}, in the USB Insertion task, the baseline VLA model (a) often misaligns the connector and continues to attempt insertion, whereas the CRAFT-enhanced model (b) continuously adjusts its motion based on contact force signals to achieve successful insertion. 
In the Flip Carton task, CRAFT similarly uses force signals to guide the direction and magnitude of applied torques, ensuring stable flipping, while the baseline fails to coordinate force and motion, resulting in task failure.

\begin{table}[t]
    \centering
    \captionsetup{justification=centering}
    \caption{Average success rates across five contact-rich manipulation tasks.}
    \label{tab:avg_success}
    \begin{tabular}{lccc}
        \toprule
        Model & w/o CRAFT & w/ CRAFT & $\Delta$(\%) \\
        \midrule
        RDT          & 22.66 & \textbf{48.32} & \textbf{25.66$\uparrow$} \\
        $\pi_0$-base & 25.32 & \textbf{60.68} & \textbf{35.36$\uparrow$} \\
        \bottomrule
    \end{tabular}
\end{table}

\textbf{CRAFT is broadly applicable to different VLA architectures.} 
CRAFT is designed as a lightweight, plug-and-play finetuning module that enhances force awareness without modifying the underlying vision-language encoders. 
This modality-agnostic design requires minimal additional supervision, making CRAFT seamlessly compatible with diverse VLA frameworks at low integration cost. 
We validate this generality by applying CRAFT to both $\pi_0$ and RDT, where the enhanced variants consistently achieve higher success rates on contact-rich manipulation tasks. 
As summarized in Tab.~\ref{tab:avg_success}, CRAFT boosts performance across different architectures, demonstrating the effectiveness of force-aware curriculum finetuning.

\begin{figure}[t]
    \centering
    \includegraphics[width=0.8\linewidth]{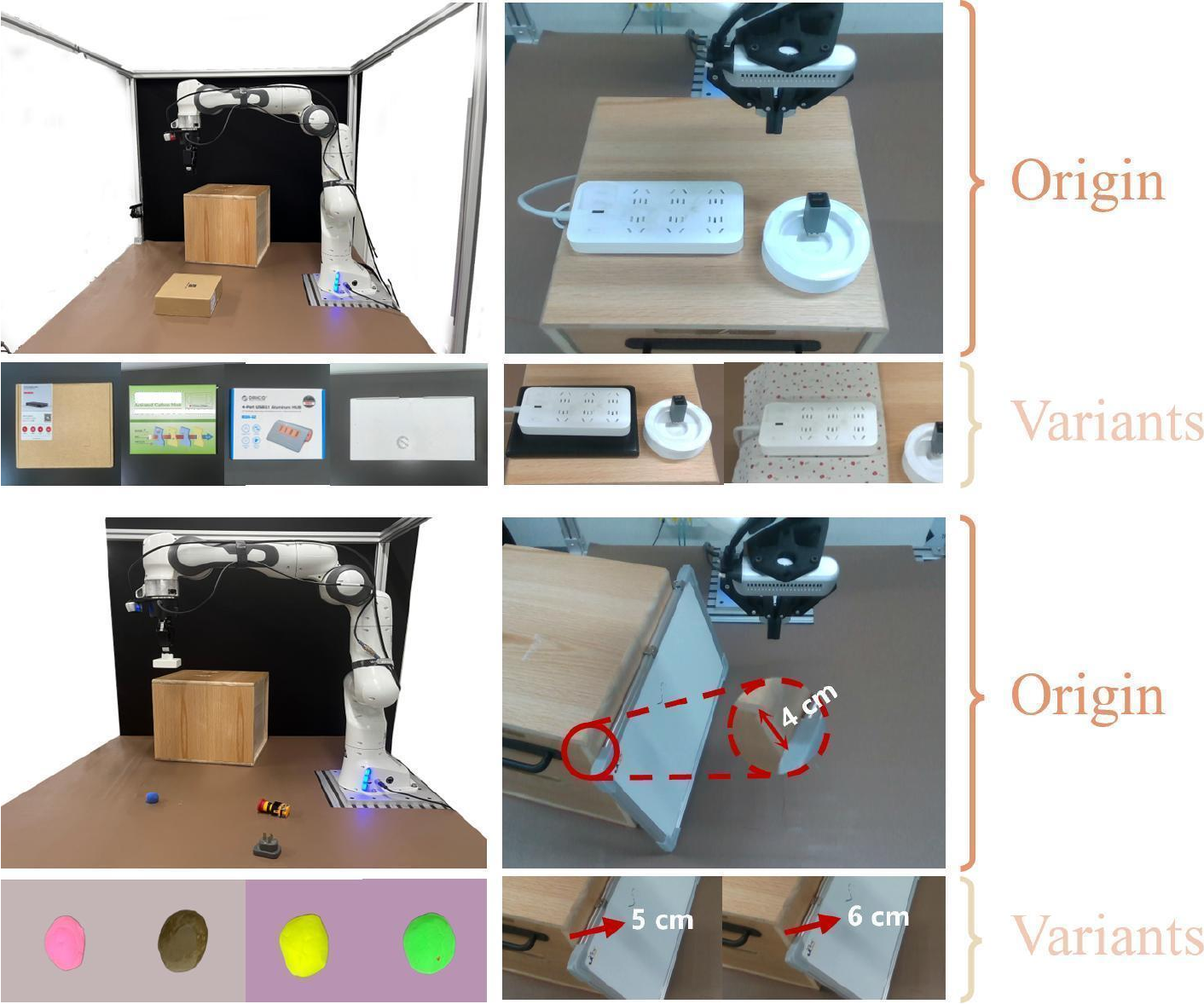}
    \caption{\textbf{Generalization experimental setups.} Top: Original experimental setups. Bottom: Variants introduced for object-level and task-level generalization.}
    \label{fig:generalization}
    \vspace{-1pt}
\end{figure}

\subsection{Generalization Studies}
We further investigate whether CRAFT enhances generalization to unseen objects and novel task variations.
Specifically, we introduce two types of Out-of-Distribution (OOD) variants:

\begin{itemize}
    \item \textit{Object-level variants}: Flip different Carton, and Rolling Plasticine with different colors.
    \item \textit{Task-level variants}: USB Insertion with socket placement height changes and visual changes, and Wipe Whiteboard with different placement angles.
\end{itemize}

As illustrated in Fig.~\ref{fig:generalization}, we evaluated the generalization performance across four contact-rich manipulation tasks—USB Insertion, Flip Carton, Wipe Whiteboard, and Rolling Plasticine—under object- and task-level OOD settings. For each variant, we conducted five evaluation trials per task, with success rates reported in Tab.~\ref{tab:gen}. The results demonstrate that CRAFT consistently improves success rates across all tasks, achieving an average increase from 22.50\% to 58.75\%, corresponding to a remarkable 36.25\% improvement. Specifically, for the task-level variant of USB Insertion, CRAFT enhances the success rate from 1/10 (10\%) to 4/10 (40\%), showing its ability to handle precise force control. Similarly, in the object-level variant of Flip Carton, performance improves from 10/20 (50\%) to 15/20 (75\%), reflecting enhanced robustness to object variations. The Wipe Whiteboard task, another task-level variant, sees a significant gain from 2/10 (20\%) to 6/10 (60\%), underscoring the effectiveness of CRAFT in tasks requiring sustained contact and force feedback. Notably, for the object-level variant of Rolling Plasticine, CRAFT achieves a substantial improvement from 2/20 (10\%) to 12/20 (60\%), highlighting its capability to adapt to visual robustness. These improvements arise from the force-aware curriculum and torque-centered representations, which enhance the model’s focus on external torques and improve its understanding of task-relevant visual and language representations. Taken together, these results validate that CRAFT significantly strengthens robustness and generalization in contact-rich manipulation tasks under OOD settings.

\begin{table}[t]
    \centering
    \captionsetup{justification=centering}
    \caption{Generalization performance on object- and task-level variants.}
    \label{tab:gen}
        \begin{tabular}{lcc}
        \toprule
        Task (Variant Type) & $\pi_0$ w/o CRAFT & $\pi_0$ w/ CRAFT \\
        \midrule
        USB Insertion (Task)           & 1/10    & \textbf{4/10} \\
        Flip Carton (Object)           & 10/20   & \textbf{15/20} \\
        Wipe Whiteboard (Task)         & 2/10    & \textbf{6/10} \\
        Rolling Plasticine (Object)    & 2/20    & \textbf{12/20} \\
        \cmidrule(lr){1-3}
        Average                        & 22.50\% & \textbf{58.75\%} \\
        \bottomrule
        \end{tabular}
\end{table}

\subsection{Ablation Studies}

To assess the contribution of each component in CRAFT, we conduct ablation experiments on the $\pi_0$-base model, focusing on three representative tasks: USB Insertion, Wipe Whiteboard, and Rolling Plasticine. Each variant is evaluated over 10 trials per task, with the reported metric being the average success rate across the three tasks. We consider the following key variants:

\begin{itemize}
    \item \textit{w/o CRAFT}: The $\pi_0$-base model.
    \item \textit{+ VIB}: Incorporates the VIB layer while using joint positions as proprioception.
    \item \textit{+ VIB \& Force}: Combines VIB with joint torque as robot state (the full CRAFT setup).
\end{itemize}

The results show that adding VIB alone substantially improves performance over the base model, highlighting the benefit of constraining visual and language embeddings to focus on proprioception information. Incorporating joint torque as the proprioception in combination with VIB further boosts success rates, confirming that torque provides richer, task-relevant feedback critical for robust contact-rich manipulation.

\begin{table}[ht]
    \centering
    \captionsetup{justification=centering}
    \caption{Ablation results.}
    \label{tab:ablation}
    \begin{tabular}{lccc}
      \toprule
      Model Variant & w/o CRAFT & + VIB & + VIB \& Force \\
      \midrule
      Avg. Success (\%)  & 20.0 & 46.7 & \textbf{56.7} \\
      \bottomrule
    \end{tabular}
\end{table} 
\section{Conclusion}
In this study, we introduced CRAFT, a force-aware curriculum fine-tuning framework designed to enhance VLA models for contact-rich manipulation. By integrating a variational information bottleneck with torque-based proprioception, CRAFT guides policies to effectively exploit force signals while simultaneously leveraging visual and language information. Experiments across five contact-rich tasks demonstrate that CRAFT improves success rates, generalizes to unseen objects and task variations, and is broadly applicable to different VLA architectures. These results highlight that incorporating a force-aware curriculum offers a lightweight yet powerful approach for achieving robust and generalizable contact-rich robotic manipulation.



\bibliographystyle{IEEEtran}
\bibliography{bibtex/bib/carft}

\end{document}